\documentclass[letterpaper, 10 pt, conference]{ieeeconf}  

\IEEEoverridecommandlockouts                              

\usepackage{amsmath}
\usepackage{amssymb}
\usepackage{graphicx}

\usepackage{amsmath}
\usepackage{amssymb}
\usepackage{algorithm}
\usepackage{algpseudocode}
\usepackage{booktabs}
\usepackage{xcolor}
\usepackage{xurl}

\title{\LARGE \bf Wheel-loader V-Cycle Automation with Deep Koopman MPC} \author{Armin Abdolmohammadi$^{1}$, Navid Mojahed$^{1}$, Dinesh Kumar$^{1}$, Bahram Ravani$^{1}$ and Shima Nazari$^{1}$ \thanks{$^{1}$The authors are with the Department of Mechanical and Aerospace Engineering, University of California, Davis, CA, USA. {\tt\small \{abdolmohammadi, nmojaed, dskumar, bravani, snazari\}@ucdavis.edu}}}

\begin{document}

\maketitle
\thispagestyle{empty}
\pagestyle{empty}

\begin{abstract}
The repeated forward–reverse maneuvers performed by wheel loaders during earthmoving operations make them well suited for automation. However, the nonlinear dynamics of articulated vehicles and complex vehicle–terrain interactions limit the effectiveness of conventional model-based approaches. This paper presents a hierarchical framework that combines long-horizon geometric planning with data-driven predictive control for autonomous wheel-loader operation. A reduced-order articulated kinematic model is used to generate the maneuver geometry, where the forward and reverse trajectories are jointly optimized through a shared intermediate state. To capture the vehicle dynamics, two data-driven deep bilinear Koopman models are learned for the forward and reverse motions using data generated from high-fidelity simulations in Algoryx Dynamics. The learned Koopman representations are subsequently incorporated into a computationally efficient model predictive control (MPC) formulation for trajectory tracking. The resulting controller operates in real time within a 50-ms execution loop. High-fidelity simulation results demonstrate that the proposed end-to-end framework enables accurate and computationally efficient execution of wheel-loader V-cycle maneuvers, providing a promising approach toward autonomous operation of articulated heavy-duty machinery.

\end{abstract}

\section{Introduction}
\label{sec:introduction}

Wheel loaders execute tightly constrained forward--reverse maneuvers during earthmoving operations. Unlike conventional single-body vehicles, they steer through articulation between two coupled bodies - linking longitudinal motion, articulation, and lateral vehicle pose. This coupling is especially critical during the short V-cycle between the pile and the truck, where the loader must approach the pile, reverse to a switchback configuration, and transition into a forward truck approach. Executing this maneuver requires both long-horizon geometric planning and real-time predictive control across changing travel directions and operating conditions.

Wheel-loader motion planning has progressed from geometric forward--reverse path construction to optimization-based formulations \cite{Gu2019longitudinal}. Reeds--Shepp-type methods established articulated forward/reverse planning \cite{Alshaer2013path}, while later work optimized the reversal location through discrete trajectory search or V-turn geometry \cite{hong2017path,aoshima2025optimizing}. These approaches do not jointly optimize the reverse and forward trajectories within a single nonlinear program in which the switchback configuration is a shared decision variable.

Predictive control for articulated heavy vehicles has progressed from analytical wheel-loader models \cite{song2022autonomous} to adaptive, learning-assisted, and direction-dependent formulations \cite{huang2024adaptive,Chen2024novel,Shi2020planning,shahirpour2025design}. However, articulated-machine MPC still relies predominantly on analytical predictive models; when learning is used, it often augments the controller through a policy or value function \cite{Maki2025motion} rather than replacing the state-transition model. Explicitly modeling transmission shifts, tire--terrain slip, and related nonlinear interactions can increase model complexity and online computational burden. Data-driven control offers an alternative when accurate analytical models are difficult to obtain \cite{zhong2025reinforcement}.

Data-driven Koopman methods provide an alternative for representing nonlinear dynamics in a form suitable for predictive control. Finite-dimensional Koopman predictors have been used to embed learned nonlinear dynamics within MPC formulations \cite{korda2018linear}. For control-affine systems, bilinear Koopman models retain state--input interactions and have been used in nonlinear and vehicle predictive control \cite{folkestad2021koopman,yu2022autonomous}. More recently, deep bilinear Koopman networks have been used to learn vehicle dynamics directly from high-fidelity data and combined with a frozen-state approximation of the bilinear terms to recover a convex real-time MPC problem \cite{abtahi2026deep}. These characteristics make articulated wheel loaders a compelling application for data-driven prediction, as articulation, wheel slip, drivetrain behavior, and terrain interaction introduce nonlinear dynamics that are difficult to capture accurately with analytical models alone.

\begin{figure*}[t]
    \centering
    \includegraphics[width=\textwidth]{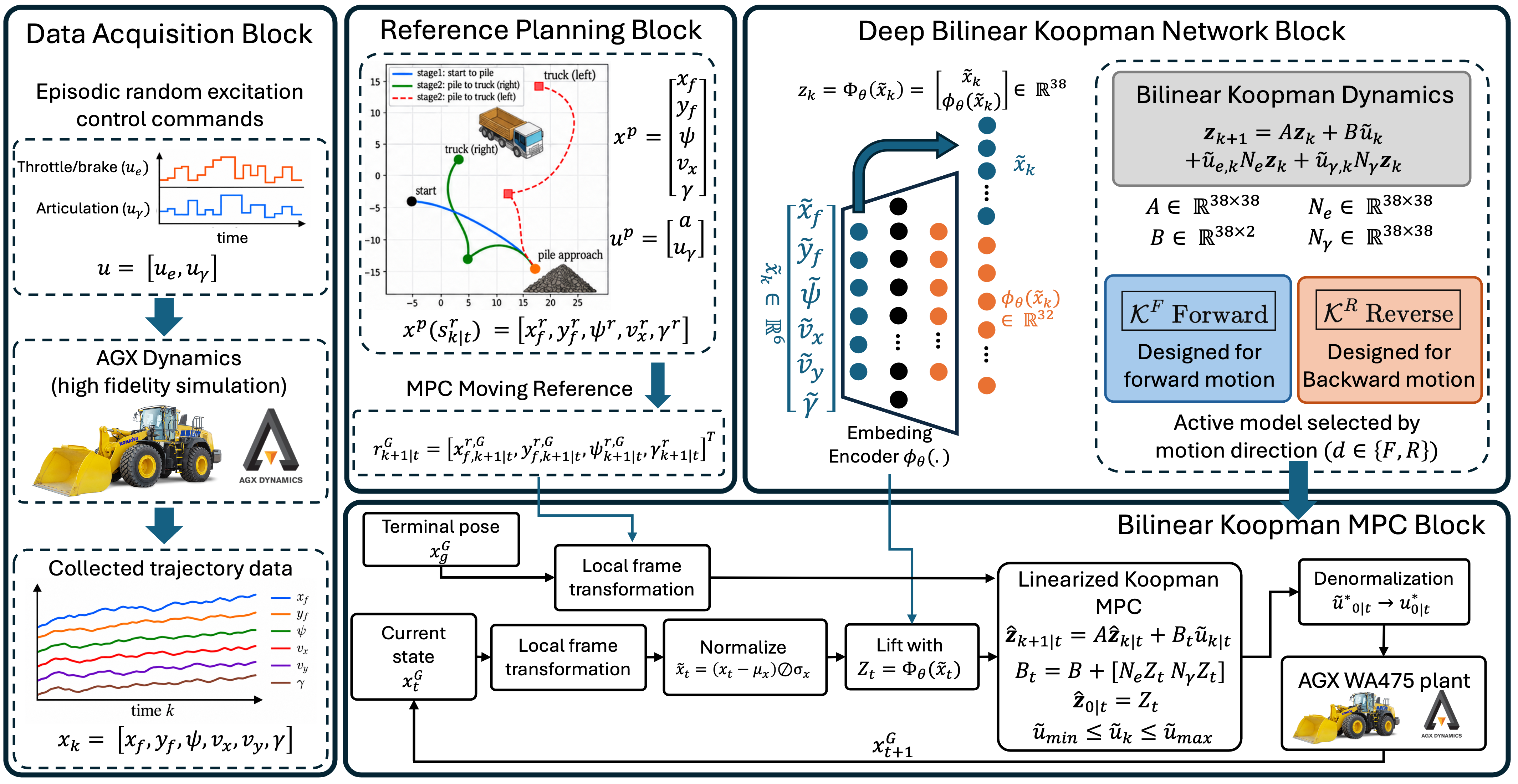}
    \caption{Overall architecture of the proposed bilinear Koopman MPC framework. Episodic AGX trajectories are used to identify separate forward and reverse deep bilinear Koopman models, while a nonlinear V-cycle planner provides the long-horizon reference. At each control update, the measured state is transformed and lifted to construct the fixed-state Koopman prediction model, and the MPC computes the control sequence applied to the AGX WA475 plant in receding-horizon linearized bilinear MPC.}
    \label{fig:overall_architecture}
\end{figure*}

This work combines long-horizon articulated-vehicle planning with direction-specific learned dynamics for real-time predictive control, as summarized in Fig.~\ref{fig:overall_architecture}. A reduced-order model generates the V-cycle reference, including a jointly optimized reverse--forward pile-to-truck maneuver, while deep bilinear Koopman models provide the predictive dynamics for short-horizon MPC. The principal contributions of this work are as follows:
\begin{itemize}
    \item Direction-specific deep bilinear Koopman models for an articulated wheel loader, trained using multi-step rollout prediction and selected based on predictive accuracy and computational cost.
    \item A nonlinear V-cycle planner that jointly optimizes the reverse and forward pile-to-truck paths through a shared continuous switchback configuration.
    \item Real-time Koopman MPC using the learned predictors, with closed-loop evaluation of the complete forward--reverse--forward V-cycle in high-fidelity simulation.
\end{itemize}

\section{System Model}
\label{sec:koopman_structure}

The target plant is a high-fidelity Komatsu WA475 model in Algoryx \cite{algoryx_agx} simulation environment, which includes an articulated multibody chassis, powertrain, compliant tires, and tire--terrain interaction.

\begin{figure}[t]
    \centering
    \includegraphics[width=0.7\columnwidth]{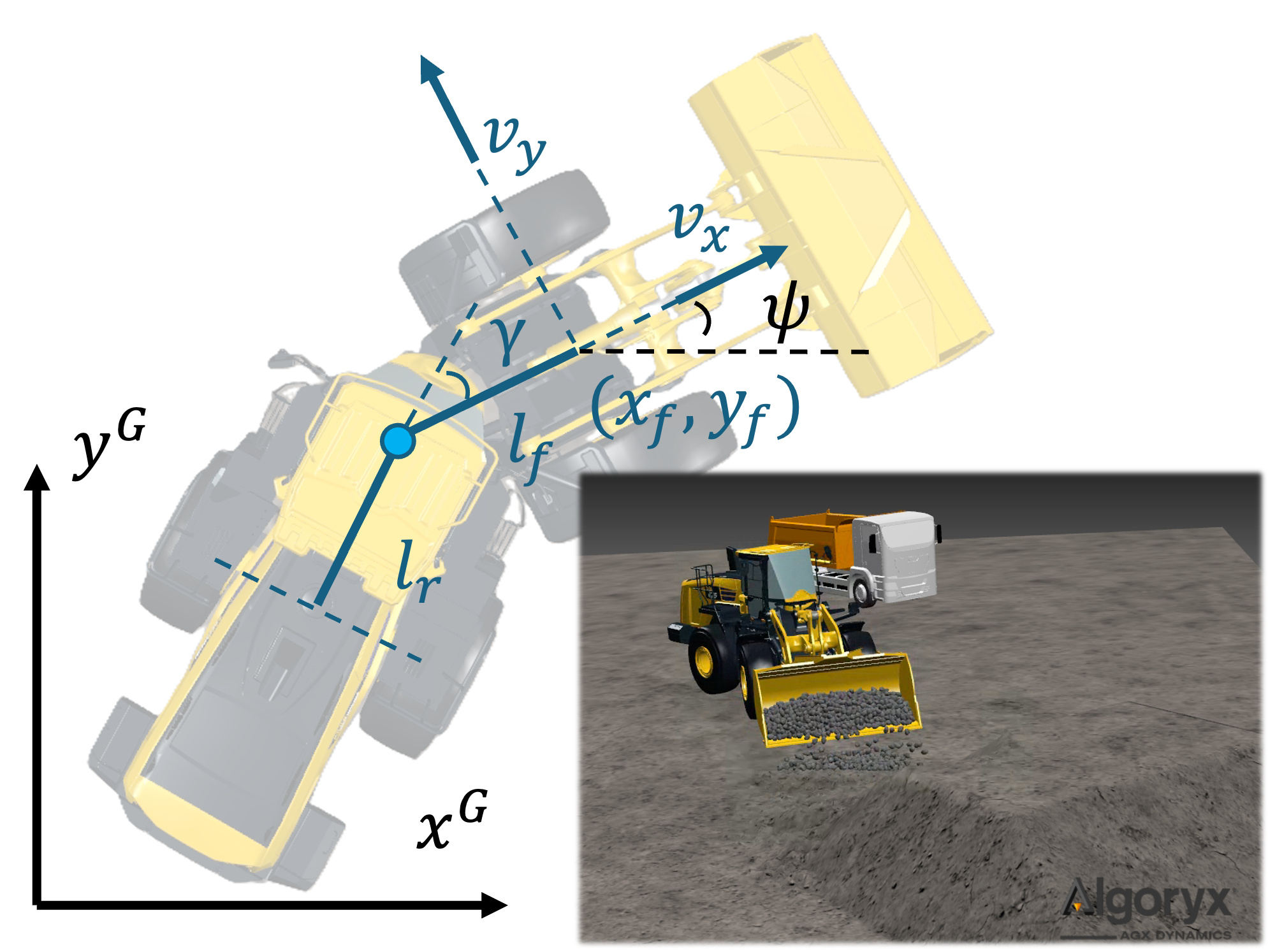}
    \caption{Articulated wheel-loader geometry, state convention, and high-fidelity Algoryx \cite{algoryx_agx} simulation environment.}
    \label{fig:loader_state}
\end{figure}

Using the front axle as the reference point, as shown in Fig.~\ref{fig:loader_state}, the state and input vectors for the wheel loader are
\begin{equation}
\mathbf{x}=
\begin{bmatrix}
x_f & y_f & \psi & v_x & v_y & \gamma
\end{bmatrix}^{\top},
\:
\mathbf{u}=
\begin{bmatrix}
u_e & u_\gamma
\end{bmatrix}^{\top},
\label{eq:koopman_state_input}
\end{equation}
where $(x_f,y_f)$ and $\psi$ are the front-axle position and heading, $v_x$ and $v_y$ are body-frame velocities, and $\gamma=\psi_r-\psi_f$ is the articulation angle. The signed effort command $u_e$ combines propulsion and braking in a single scalar input, with $u_e>0$ corresponding to positive throttle and $u_e<0$ corresponding to braking. $u_\gamma=\dot{\gamma}_{\mathrm{cmd}}=\dot{\gamma}$ is assumed under perfect articulation-rate actuation. The transmission gear is selected outside the continuous vehicle model. Forward and reverse operations are therefore treated as separate modes, avoiding the introduction of a discrete gear variable and the resulting mixed-integer predictive-control formulation.

We define the reduced planning state and input as
\begin{equation}
\mathbf{x}^p
=
\begin{bmatrix}
x_f &
y_f &
\psi &
v_x &
\gamma
\end{bmatrix}^{\top},
\qquad
\mathbf{u}^p
=
\begin{bmatrix}
a &
u_\gamma
\end{bmatrix}^{\top},
\label{eq:planner_state_input}
\end{equation}
where the superscript $p$ denotes variables used by the motion planner. Here, $a$ is the longitudinal acceleration command and $u_\gamma$ is the commanded articulation rate. The planning model leaves out the lateral velocity $v_y$ by assuming $v_y=0$, and replaces the effort command $u_e$ with longitudinal acceleration $a$. These assumptions simplify the vehicle representation and make it well suited for long-horizon trajectory optimization. Using the same front-axle reference convention, the articulated kinematic model is given by \cite{Ridley2003load,Chen2024novel}
\begin{equation}
\dot{\mathbf{x}^p}
=
\left[
v_x\cos\psi,\:
v_x\sin\psi,\:
-\frac{v_x\sin\gamma+l_r u_\gamma}{l_f\cos\gamma+l_r},\:
a,\:
u_\gamma
\right]^\top,
\label{eq:planner_dynamics}
\end{equation}
where $l_f$ and $l_r$ are the distances from the articulation joint to the front and rear axles, respectively.

\subsection{Deep Bilinear Koopman Model}
\label{subsec:bilinear_koopman}

Consider a discrete-time nonlinear system as $\mathbf{x}_{k+1}
=F(\mathbf{x}_k,\mathbf{u}_k)$ where $k$ denotes the time index, $\mathbf{x}\in \mathbb{R}^{n_x}$ and $\mathbf{u}\in \mathbb{R}^{n_u}$. Koopman operator based methods represent nonlinear dynamics through the linear evolution of observable functions in a higher-dimensional function space \cite{korda2018linear,brunton2022modern}. Traditional approaches rely on a manual selection of observable functions, which limits the linearization accuracy to the collective expressiveness of the observable library. 

In the deep Koopman approach, on the other hand, a neural encoder $\phi_\theta: \mathbb{R}^{n_x} \rightarrow \mathbb{R}^{n_\phi}$ is used to learn observable functions directly from data \cite{lusch2018deep}. In this work we first normalize state and input variables as
\begin{equation}
\tilde{x}_i
=
\frac{x_i-\mu_{x_i}}{\sigma_{x_i}},
\qquad
\tilde{u}_i
=
\frac{u_i-\mu_{u_i}}{\sigma_{u_i}},
\label{eq:normalization}
\end{equation}
where the normalization mean $\mu$ and standard deviation $\sigma$ are computed from the corresponding training dataset only. The lifted state is formed by concatenating the normalized physical state with the learned observables from the deep network (see Fig. \ref{fig:overall_architecture}) as:
\begin{equation}
\mathbf{z}_k
=
\Phi_\theta(\tilde{\mathbf{x}}_k)
=
\begin{bmatrix}
\tilde{\mathbf{x}}_k\\
\phi_\theta(\tilde{\mathbf{x}}_k)
\end{bmatrix}
\in\mathbb{R}^{n_z},
\qquad
n_z=n_x+n_\phi.
\label{eq:lifted_state}
\end{equation}
The normalized physical state is recovered through a linear projection
\begin{equation}
\tilde{\mathbf{x}}_k
=
C\mathbf{z}_k,
\qquad
C=
\begin{bmatrix}
I_{n_x} & 0
\end{bmatrix}.
\label{eq:koopman_projection}
\end{equation}
For a predicted lifted state $\hat{\mathbf z}_k$, the corresponding normalized
physical-state prediction is therefore $\hat{\tilde{\mathbf{x}}}_k=C\hat{\mathbf z}_k$, followed by a denormalization to recover the physical units.

The controlled lifted dynamics are represented using a bilinear Koopman model as \cite{abtahi2026deep,bruder2021advantages}:
\begin{equation}
\mathbf{z}_{k+1}
=
A\mathbf{z}_k
+
B\tilde{\mathbf{u}}_k
+
\tilde u_{e,k}N_e\mathbf{z}_k
+
\tilde u_{\gamma,k}N_\gamma\mathbf{z}_k.
\label{eq:bilinear_koopman}
\end{equation}
Here, $A,N_e,N_\gamma\in\mathbb{R}^{n_z\times n_z}$ and $B\in\mathbb{R}^{n_z\times n_u}$. The matrices $N_e$ and $N_\gamma$ capture the lifted-state interactions with longitudinal effort and articulation rate, respectively.

\subsection{Loss function parametrization}
\label{subsec:koopman_losses}

The neural encoder and Koopman matrices are optimized jointly from trajectory sequences using a composite objective that promotes \textit{one-step consistency}, \textit{recursive multi-step prediction}, and \textit{stable, regularized lifted dynamics}. Let $N_{\mathrm{tr}}$ denote the number of steps in a training trajectory. For \textit{one-step prediction}, the measured state at each time index is lifted and propagated through \eqref{eq:bilinear_koopman}. Denoting the resulting prediction by $\hat{\mathbf z}_{k+1}^{(1)}$, the one-step lifted prediction loss is
\begin{equation}
\mathcal{L}_{z,1}
=
\frac{1}{N_{\mathrm{tr}}n_z}
\sum_{k=0}^{N_{\mathrm{tr}}-1}
\left\|
\hat{\mathbf z}_{k+1}^{(1)}
-
\Phi_\theta(\tilde{\mathbf x}_{k+1})
\right\|_2^2.
\label{eq:one_step_lifted_loss}
\end{equation}

For recursive multi-step prediction, only the initial state of each training
sequence is lifted, $\hat{\mathbf z}_0 = \Phi_\theta(\tilde{\mathbf x}_0)$, after which the lifted state is propagated recursively using
\eqref{eq:bilinear_koopman} without intermediate re-encoding. The
corresponding lifted-space rollout loss is
\begin{equation}
\mathcal{L}_{z,m}
=
\frac{1}{N_{\mathrm{tr}}n_z}
\sum_{k=1}^{N_{\mathrm{tr}}}
\left\|
\hat{\mathbf z}_{k}
-
\Phi_\theta(\tilde{\mathbf x}_{k})
\right\|_2^2.
\label{eq:multi_step_lifted_loss}
\end{equation}

For a bilinear model, the homogeneous lifted-state transition depends on the
input according to
\begin{equation}
A_{\mathrm{eff}}(\tilde{\mathbf u})
=
A
+
\tilde u_eN_e
+
\tilde u_\gamma N_\gamma.
\label{eq:effective_A}
\end{equation}
A spectral penalty is evaluated over a finite physical-input grid $\mathcal U_g$ spanning the modeled operating region. Each grid point is standardized using \eqref{eq:normalization} before evaluating $A_{\mathrm{eff}}$. Denoting the corresponding normalized grid by $\tilde{\mathcal U}_g$, the penalty is
\begin{equation}
\mathcal{L}_{\rho}
=
\frac{1}{|\tilde{\mathcal U}_g|n_z}
\sum_{\tilde{\mathbf u}\in\tilde{\mathcal U}_g}
\sum_{j=1}^{n_z}
\left[
\max
\left(
0,
\left|
\lambda_j
\left(
A_{\mathrm{eff}}(\tilde{\mathbf u})
\right)
\right|
-1
\right)
\right]^2.
\label{eq:spectral_loss}
\end{equation}
This term discourages unstable lifted dynamics over the sampled input region but does not constitute a formal stability guarantee.

Parameter regularization is applied separately to the base model and the bilinear coupling matrices. Let $\Theta_\phi$ denote the collection of trainable encoder parameter tensors. The regularization terms are
\begin{equation}
\begin{aligned}
\mathcal L_{\mathrm{reg}}
&=
\frac{\|A\|_F^2}{n_z^2}
+
\frac{\|B\|_F^2}{n_zn_u}
+
\sum_{W\in\Theta_\phi}
\frac{\|W\|_F^2}{|W|},
\\
\mathcal L_{\mathrm{bil}}
&=
\frac{\|N_e\|_F^2}{n_z^2}
+
\frac{\|N_\gamma\|_F^2}{n_z^2}.
\end{aligned}
\label{eq:koopman_regularization}
\end{equation}
where $\|\cdot\|_F$ denotes the Frobenius norm and $|W|$ is the number of
scalar parameters in $W$.

The loss terms are combined as
\begin{equation}
\begin{aligned}
\mathcal L ={}& w_{z,1}\mathcal{L}_{z,1} + w_{z,m}\mathcal{L}_{z,m} +
w_{\rho}\mathcal{L}_{\rho}
\\
&+ w_{\mathrm{reg}}\mathcal{L}_{\mathrm{reg}} + w_{\mathrm{bil}}\mathcal{L}_{\mathrm{bil}}.
\end{aligned}
\label{eq:koopman_training_objective}
\end{equation}
The encoder and Koopman matrices are optimized jointly through backpropagation.

\subsection{Data Acquisition and Preprocessing}
\label{subsec:data_acquisition}

Forward- and reverse-motion datasets are collected independently from the AGX WA475 plant, with $2000$ successful stop-to-stop episodes retained for each direction. Each episode comprises launch, randomized excitation, and controlled braking to rest. During the $8$--$15$~s excitation phase, the input pair $(u_e,u_\gamma)$ is held piecewise constant for random intervals of $0.20$--$0.80$~s. The effort command spans $u_e\in[-1,1]$, while the articulation-rate command spans $u_\gamma\in[-0.30,0.30]$~rad/s. The states and commands are recorded at $\Delta t=0.05$~s.

For each direction, the $2000$ retained episodes are split at the episode level into $1400$ training, $300$ validation, and $300$ test episodes to prevent portions of the same trajectory from appearing across datasets. Training sequences of $5$~s are then extracted with a $0.5$-s stride, yielding $101$ state samples and $100$ input samples per sequence.

To remove dependence on absolute position and heading, each sequence is expressed in a local frame attached to its initial front-axle pose. Defining $\Delta x_{f,k}^{G}=x_{f,k}^{G}-x_{f,0}^{G}$ and $\Delta y_{f,k}^{G}=y_{f,k}^{G}-y_{f,0}^{G}$,
\begin{equation}
\begin{aligned}
x_{f,k}
&=
\cos\psi_0^G\,\Delta x_{f,k}^{G}
+
\sin\psi_0^G\,\Delta y_{f,k}^{G},\\
y_{f,k}
&=
-\sin\psi_0^G\,\Delta x_{f,k}^{G}
+
\cos\psi_0^G\,\Delta y_{f,k}^{G},\\
\psi_k
&=
\psi_k^G-\psi_0^G.
\end{aligned}
\label{eq:training_local_frame}
\end{equation}
Thus, every sequence begins from $x_{f,0}=y_{f,0}=\psi_0=0$. The global heading is unwrapped before transformation, while $v_x$, $v_y$, and $\gamma$ require no additional transformation because they are already body-relative quantities.

State and input normalization statistics are computed from the training split of each direction and applied unchanged to its validation and test sets. The forward and reverse models therefore retain independent normalization throughout training and control.

\subsection{Training Procedure}
\label{subsec:training_procedure}

The encoder and Koopman matrices are jointly optimized by minimizing \eqref{eq:koopman_training_objective} using mini-batch Adam, with forward and reverse models trained independently. Hyperparameters are selected separately for each direction using Optuna with a TPE sampler \cite{akiba2019optuna,bergstra2011algorithms}, varying the encoder architecture, latent dimension, batch size, learning rate, and spectral-penalty weight. Trials are ranked using a validation criterion combining recursive physical-state prediction error, terminal-position RMSE, and excessive spectral growth. Within each trial, the best checkpoint is selected from the validation rollout performance, including physical- and lifted-state prediction errors and the spectral penalty. Validation is performed every $10$ epochs; the learning rate is halved after five checks without improvement, early stopping is applied after $12$ such checks, and training is limited to $300$ epochs.

\section{V-Cycle Motion Planning}
\label{sec:vcycle_planning}

A wheel-loader loading cycle requires coordinated motion between a material pile and a haul truck. In this work, the vehicle first travels from its initial configuration to a prescribed pile-approach pose and then executes a pile-to-truck transfer; excavation is treated separately as done by \cite{Abdolmohammadi2025Optimal,Abdolmohammadi2025Data}. The initial-to-pile stage is solved as a single forward optimization, while the pile-to-truck maneuver is formulated as a coupled reverse--forward problem with an optimized switchback configuration. The truck-approach side may be prescribed or selected by solving the same nonlinear program in parallel from left- and right-approach initializations. The subsequent truck-to-initial return follows the same planning and execution procedure as the initial-to-pile maneuver and is therefore omitted to avoid redundancy.

The planners described in this section use the reduced-order articulated kinematic model defined in \eqref{eq:planner_state_input}--\eqref{eq:planner_dynamics}. 

\subsection{Initial-to-Pile Planning}
\label{subsec:pile_planning}
The initial-to-pile maneuver is formulated as:
\begin{equation}
\begin{aligned}
\min_{\mathbf{x}^p,\mathbf{u}^p}\:
J_{\mathrm{I}}
&=
\|\mathbf{e}_{P,N_P}\|_{Q_P}^{2}
+\sum_{k=0}^{N_P-1}\|\mathbf{u}_k^p\|_{R_P}^{2}
\\
\mathrm{s.t.}\quad
&\mathbf{x}_0^p=\mathbf{x}_{\mathrm{start}}^p,
\\
&\mathbf{x}_{k+1}^p
=F_{\Delta t_P}(\mathbf{x}_k^p,\mathbf{u}_k^p),
\\
&0\leq v_{x,k}\leq v_{x,\max},
\qquad |\gamma_k|\leq\gamma_{\max},
\\
&|u_{\gamma,k}|\leq u_{\gamma,\max},
\qquad a_{\min}\leq a_k\leq a_{\max},
\\
&v_{x,N_P}=v_{x,g}.
\end{aligned}
\label{eq:pile_planner}
\end{equation}
Where $N_P$ is the problem horizon, dynamics and input constraints apply for $k=0,\ldots,N_P-1$, the state bounds apply for $k=0,\ldots,N_P$ and the model is discretized using fourth-order Runge--Kutta integration as $\mathbf{x}^{p}_{k+1}=F_{\Delta t_P}(\mathbf{x}^{p}_k,\mathbf{u}^{p}_k)$. Here $Q_P\succeq0$ and $R_P\succeq0$ are diagonal weighting matrices. The terminal configuration $(x_g,y_g,\psi_g,\gamma_g)$ is specified in front of the pile, while the terminal longitudinal velocity is imposed separately through the hard constraint $v_{x,N_P}=v_{x,g}$ to enforce a full stop. The corresponding terminal configuration error in \eqref{eq:pile_planner} is defined as
\begin{equation}
\begin{aligned}
\mathbf{e}_{P,N_P}
=\big[&
x_{f,N_P}-x_g,\;
y_{f,N_P}-y_g,\\
&
\delta_\pi(\psi_{N_P},\psi_g),\;
\gamma_{N_P}-\gamma_g
\big]^{\top}.
\end{aligned}
\label{eq:pile_terminal_error}
\end{equation}
where $\delta_\pi(\alpha,\beta)=\operatorname{atan2}(\sin(\alpha-\beta),\cos(\alpha-\beta))$ denotes the wrapped angular difference.

\subsection{Pile-to-Truck Dual-Stage Planning}
\label{subsec:truck_planning}

The pile-to-truck maneuver is formulated as a coupled reverse--forward optimization over $N_R$ and $N_F$ intervals. Both stages use the same kinematic model, with travel direction imposed through signed longitudinal-velocity bounds, while the intermediate switchback configuration is optimized jointly with both trajectories. For each stage, define $\boldsymbol{\eta}_k^d=[\gamma_k^d,\;v_{x,k}^d-v_{\mathrm{ref}}^d]^\top$ and $\Delta\mathbf{u}_k^{p,d}=\mathbf{u}_{k+1}^{p,d}-\mathbf{u}_k^{p,d}$. The stage cost is
\begin{equation}
\begin{aligned}
J_d
={}& \sum_{k=0}^{N_d-1} \left( \|\boldsymbol{\eta}_k^d\|_{Q_d}^{2} + \|\mathbf{u}_k^{p,d}\|_{R_d}^{2} + w_{\mathrm{ls}}
\frac{(u_{\gamma,k}^{d})^2}
{(v_{x,k}^{d})^2+\epsilon_{\mathrm{ls}}^2}\right)
\\
&+ \sum_{k=0}^{N_d-2} \|\Delta\mathbf{u}_k^{p,d}\|_{R_\Delta}^{2}, \qquad d\in\{R,F\}.
\end{aligned}
\label{eq:dual_stage_cost}
\end{equation}
where $\epsilon_{\mathrm{ls}}>0$ is a small regularization constant that prevents singular behavior near zero longitudinal velocity. Here $v_{\mathrm{ref}}^R<0$ and $v_{\mathrm{ref}}^F>0$ are the nominal reverse and forward velocities. The cost penalizes articulation, speed error, control effort, and variation, while the low-speed term discourages large articulation-rate commands near a stop.

The terminal truck-pose error is defined as
\begin{equation}
\begin{aligned}
\mathbf{e}_g
=\big[&
x_{f,N_F}^{F}-x_g,\;
y_{f,N_F}^{F}-y_g,\\
&
\delta_\pi(\psi_{N_F}^{F},\psi_g),\;
\gamma_{N_F}^{F}-\gamma_g
\big]^{\top},
\end{aligned}
\label{eq:truck_terminal_error}
\end{equation}
with terminal cost $J_g = \|\mathbf{e}_g\|_{Q_g}^{2}$, where $Q_g$ is diagonal.

The switchback and truck-goal positions are $\mathbf p_s=[x_{f,N_R}^{R},\,y_{f,N_R}^{R}]^\top$ and $\mathbf p_g=[x_g,\,y_g]^\top$, respectively.
The desired heading at the switchback is taken as the direction from the
switchback toward the truck, $\psi_s^\star = \angle\left(\mathbf p_g-\mathbf p_s\right)$, where $\angle(\cdot)$ is the planar angle of a two-dimensional vector.

To encourage sufficient reverse displacement before transitioning to the forward stage, define the unit vector along the initial reverse direction as $\mathbf d_0^R=[-\cos\psi_0^R,\,-\sin\psi_0^R]^\top$
 and the reverse progress as
\begin{equation}
d_R
=
\left(
\mathbf p_s-\mathbf p_0^R
\right)^\top
\mathbf d_0^R,
\;
\mathbf p_0^R=[x_{f,0}^{R},\,y_{f,0}^{R}]^\top.
\label{eq:reverse_progress}
\end{equation}

The switchback cost is then
\begin{equation}
J_s
=
w_{\gamma,s}
\left(\gamma_{N_R}^{R}\right)^2
+
w_{\psi,s}
\left[
\delta_\pi(\psi_{N_R}^{R},\psi_s^\star)
\right]^2
-
w_{\mathrm{prog}}d_R .
\label{eq:switchback_cost}
\end{equation}

The reverse stage is initialized from the measured vehicle state at the end of the initial-to-pile maneuver. The coupled reverse--forward optimization is
\begin{equation}
\begin{aligned}
\min_{\substack{
\mathbf{x}^{p,R},\,\mathbf{u}^{p,R},\\
\mathbf{x}^{p,F},\,\mathbf{u}^{p,F}
}}
&
 J_{\mathrm{II}} = J_R +J_F + J_s + J_g \\
\mathrm{s.t.}\quad 
&\mathbf{x}^{p,R}_{0} =
\mathbf{x}^{p}_{\mathrm{meas}},\\
& \mathbf{x}^{p,R}_{k+1} = F_{\Delta t_R}\left(\mathbf{x}^{p,R}_k,\mathbf{u}^{p,R}_k\right),\: k=0,\ldots,N_R-1, \\
& v_{x,\min} \leq v_{x,k}^{R} \leq 0, \: k=0,\ldots,N_R, \\
& \mathbf{x}^{p,F}_{k+1} = F_{\Delta t_F} \left( \mathbf{x}^{p,F}_k, \mathbf{u}^{p,F}_k \right), \: k=0,\ldots,N_F-1, \\
& 0 \leq v_{x,k}^{F} \leq v_{x,\max}, \: k=0,\ldots,N_F, \\
& \mathbf{x}^{p,F}_{0} = \mathbf{x}^{p,R}_{N_R}, \\
& v_{x,N_R}^{R} = v_{x,0}^{F} = 0, \\
& v_{x,N_F}^{F} = v_{x,g}, \\
& |\gamma_k^d| \leq \gamma_{\max}, \: d\in\{R,F\}, \\
& |u_{\gamma,k}^{d}| \leq u_{\gamma,\max}, \: a_{\min} \leq a_k^d \leq a_{\max}, \:d\in\{R,F\}.
\end{aligned}
\label{eq:dual_stage_planner}
\end{equation}
Each optimal control problem is solved once at the beginning of the corresponding movement phase using IPOPT \cite{wachter2006implementation}. The initial-to-pile problem is solved from the measured vehicle state at the start of the cycle, while the coupled reverse--forward pile-to-truck problem is solved after the loader reaches the pile-approach configuration. The resulting piecewise trajectory provides the long-horizon reference for the Koopman MPC in Section~\ref{sec:koopman_mpc}.

\begin{figure*}[t]
    \centering
    \includegraphics[width=0.9\textwidth]{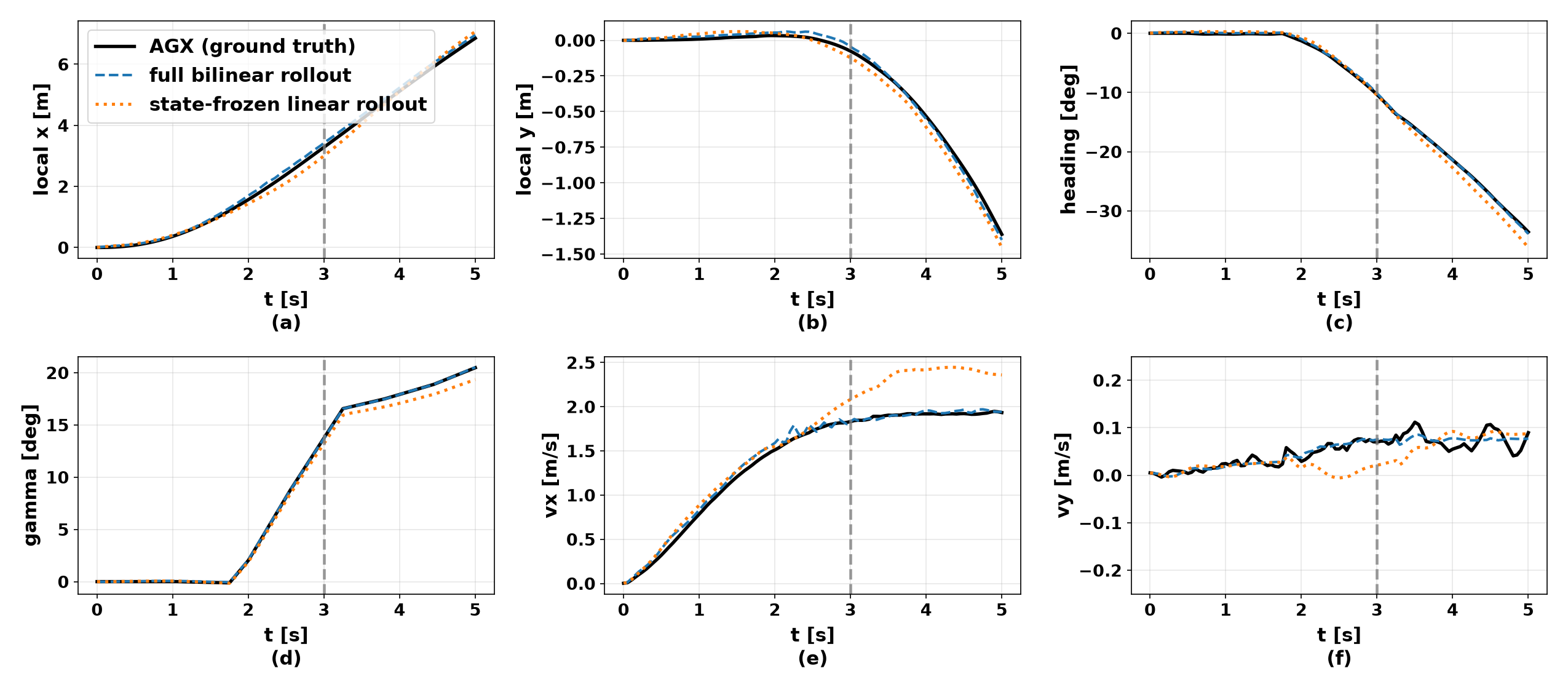}
    \caption{Example forward-motion open-loop prediction over a 5-s held-out AGX trajectory. The full bilinear Koopman model and the fixed-state linearized approximation are compared with the AGX ground truth for all six physical states. The vertical dashed line denotes the 3-s prediction horizon used by the deployed MPC.}
    \label{fig:koopman_openloop}
\end{figure*}

\section{Bilinear Koopman Model Predictive Control}
\label{sec:koopman_mpc}

The long-horizon V-cycle trajectory generated in Section~\ref{sec:vcycle_planning} is tracked using a short-horizon Koopman MPC. As illustrated in Fig.~\ref{fig:overall_architecture}, Stages~1 and~3 use the forward model $\mathcal K^F$, while Stage~2 uses the independently identified reverse model $\mathcal K^R$. For compactness, the direction superscript is omitted in the following derivation.

\subsection{Frozen-State Linearization of Bilinear Model}
\label{subsec:koopman_linearization}

Direct use of the bilinear Koopman model in \eqref{eq:bilinear_koopman} introduces products between the control input and lifted state. To obtain an affine prediction model over each MPC horizon, the lifted state in the bilinear terms is frozen at the current lifted measured state \cite{abtahi2026deep}. At control update $t$, $\mathbf Z_t=\Phi_\theta(\tilde{\mathbf x}_t)$, and the effective input matrix is
\begin{equation}
B_t
=
B+
\begin{bmatrix}
N_e\mathbf Z_t &
N_\gamma\mathbf Z_t
\end{bmatrix}.
\label{eq:frozen_input_matrix}
\end{equation}
The prediction model is then
\begin{equation}
\hat{\mathbf z}_{k+1|t}
=
A\hat{\mathbf z}_{k|t}
+
B_t\tilde{\mathbf u}_{k|t},
\qquad
\hat{\mathbf z}_{0|t}
=
\mathbf Z_t .
\label{eq:linearized_koopman_mpc}
\end{equation}

The matrix $B_t$ is recomputed at every control update and remains fixed over the corresponding prediction horizon.

\subsection{Moving Path Reference}
\label{subsec:moving_reference}

Each MPC problem is expressed in the vehicle-local frame defined by \eqref{eq:training_local_frame}, with the current front-axle pose as the origin. The stage goal and planned path are transformed into the same frame.

Let $s_i$ denote the cumulative arclength of path sample $i$. On the first control update of each stage, the anchor is selected from the complete stage path; thereafter, the search is restricted to $\mathcal I_t=\{i:s_{t-1}^a\leq s_i\leq s_{t-1}^a+c_wN\Delta s_{\mathrm{ref}}\}$ to prevent backward jumps in arclength. The anchor is
\begin{equation}
i_t
=
\arg\min_{i\in\mathcal I_t}
\|\mathbf p_t^{G}-\mathbf p_i^{G}\|_2^2,
\qquad
s_t^a=s_{i_t}.
\label{eq:path_anchor}
\end{equation}

The $N$ reference points are sampled ahead of the anchor according to
\begin{equation}
s_{k+1|t}^{r}
=
\min\!\left(
s_t^a+(k+1)\Delta s_{\mathrm{ref}},
s_{\mathrm{end}}
\right),
\:
k=0,\ldots,N-1.
\label{eq:moving_reference_arclength}
\end{equation}
Position, heading, and articulation are interpolated at these arclengths, with heading unwrapped before interpolation to avoid $\pm\pi$ discontinuities. After transformation to the current local frame, the running reference is 
\begin{equation} \mathbf r_{k+1|t} = \begin{bmatrix} x_{k+1|t}^{r} & y_{k+1|t}^{r} & \psi_{k+1|t}^{r} & \gamma_{k+1|t}^{r} \end{bmatrix}^{\top}. \label{eq:running_reference} \end{equation} The running reference excludes $v_x$ and $v_y$ since the kinematic planner is used primarily to provide long-horizon geometric guidance.

\subsection{Path Following MPC Formulation}
\label{subsec:mpc_formulation}
Let $\hat{\mathbf x}_{k|t}$ denote the physical state decoded from $\hat{\mathbf z}_{k|t}$. The moving-reference tracking error then becomes
\begin{equation}
\begin{aligned}
\mathbf e_{k+1|t}^{r}
=\big[&
\hat x_{f,k+1|t}-x_{k+1|t}^{r},\;
\hat y_{f,k+1|t}-y_{k+1|t}^{r},\\
&
\hat\psi_{k+1|t}-\psi_{k+1|t}^{r},\;
\hat\gamma_{k+1|t}-\gamma_{k+1|t}^{r}
\big]^\top.
\end{aligned}
\label{eq:mpc_running_error}
\end{equation}
where it is evaluated for $k=0,\ldots,N-1$. The fixed stage target is transformed from the global frame into the current vehicle-local frame before each MPC solve. Its local representation is $\mathbf{x}_g =[x_g,\;y_g,\;\psi_g,\;v_{x,g},\;v_{y,g},\;\gamma_g]^\top$.
Because each stage terminates at rest, the desired terminal velocities are set to $v_{x,g}=v_{y,g}=0$. The corresponding terminal error is
\begin{equation}
\begin{aligned}
\mathbf e_{N|t}^{g}
=\big[&
\hat x_{f,N|t}-x_g,\;
\hat y_{f,N|t}-y_g,\;
\hat\psi_{N|t}-\psi_g,\\
&
\hat v_{x,N|t}-v_{x,g},\;
\hat v_{y,N|t}-v_{y,g},\;
\hat\gamma_{N|t}-\gamma_g
\big]^\top.
\end{aligned}
\label{eq:mpc_terminal_error}
\end{equation}

As defined in \eqref{eq:normalization}, the control input used by the learned Koopman model is normalized. Accordingly, $\tilde{\mathbf u}_{k|t}$ is used as the MPC optimization variable, while the corresponding physical command is recovered as $\mathbf u_{k|t} = \boldsymbol{\mu}_u + D_u\tilde{\mathbf u}_{k|t}$, where $D_u=\operatorname{diag}(\boldsymbol{\sigma}_u)$. The physical input $\mathbf u_{k|t}$ is used in the control penalties and input constraints, with $\Delta\mathbf u_{0|t}=\mathbf u_{0|t}-\mathbf u_{t-1}$ and $\Delta\mathbf u_{k|t}=\mathbf u_{k|t}-\mathbf u_{k-1|t}$ for $k\geq1$.

With these quantities defined, diagonal weighting matrices $Q_r$, $Q_g$, $R$, and $R_\Delta$ are used to penalize moving-reference tracking error, terminal-goal error, control effort, and control variation, respectively. The finite-horizon MPC problem then becomes:
\begin{equation}
\begin{aligned}
\min_{\tilde{\mathbf u}_{0:N-1|t}} \quad
&
\sum_{k=0}^{N-1} \Big( \|\mathbf e^r_{k+1|t}\|_{Q_r}^2 + \|\mathbf u_{k|t}\|_R^2 +\|\Delta\mathbf u_{k|t}\|_{R_\Delta}^2\Big)\\
&\quad
+
\|\mathbf e^g_{N|t}\|_{Q_g}^2
\\
\mathrm{s.t.}\quad
&\hat{\mathbf z}_{0|t}=\mathbf Z_t,
\\
&\hat{\mathbf z}_{k+1|t}
=
A\hat{\mathbf z}_{k|t}
+
B_t\tilde{\mathbf u}_{k|t},
\\
&\mathbf u_{k|t}
=
\boldsymbol{\mu}_u
+
D_u\tilde{\mathbf u}_{k|t},
\\
&\mathbf u_{\min}
\leq
\mathbf u_{k|t}
\leq
\mathbf u_{\max},
\\
&\gamma_{\min}
\leq
\hat{\gamma}_{k+1|t}
\leq
\gamma_{\max}.
\end{aligned}
\label{eq:koopman_mpc_problem}
\end{equation}
The physical input bounds constrain effort and articulation rate, while input variation is penalized rather than imposed as a hard constraint. The lifted states are recursively substituted using \eqref{eq:linearized_koopman_mpc}, leaving only the normalized control sequence as the optimization variable. The previous optimal sequence is shifted by one step to warm-start the subsequent solve.

\subsection{Stage Completion and Handover}
\label{subsec:stage_handover}
\begin{table}[t]
\centering
\caption{Stage-specific Koopman MPC parameters.}
\label{tab:mpc_parameters}
\footnotesize
\setlength{\tabcolsep}{2.8pt}
\begin{tabular}{lccc}
\hline
\textbf{Parameter} & \textbf{Stage 1} & \textbf{Stage 2} & \textbf{Stage 3} \\
\hline
Koopman model & Forward & Reverse & Forward \\
$\Delta s_{\mathrm{ref}}$ [m] & 0.25 & 0.15 & 0.11 \\
$q_\gamma/10^3$ & 1 & 50 & 100 \\
$Q_g/10^2$ &
\shortstack{$\operatorname{diag}(0.01,0.01,1,$\\
$0.01,0.01,1)$} &
\shortstack{$\operatorname{diag}(1,10,100,$\\
$0.01,0.01,500)$} &
\shortstack{$\operatorname{diag}(1,1,10,$\\
$0.01,0.01,10)$} \\
\hline
$\varepsilon_p$ [m] & 2.0 & 2.0 & 2.0 \\
$\varepsilon_\psi$ [deg] & 15 & 15 & 15 \\
$\varepsilon_v$ [m/s] & 0.5 & 0.5 & 0.5 \\
$\varepsilon_\gamma$ [deg] & 5 & 10 & 10 \\
\hline
\end{tabular}
\end{table}
Because the terminal cost provides only soft convergence, stage completion is evaluated separately from the measured global vehicle state. Let $\mathbf p_t$ and $\mathbf p_g$ denote the measured and target front-axle positions, respectively.

A stage is completed when
\begin{equation}
\begin{aligned}
\|\mathbf p_t-\mathbf p_g\|_2
&\leq \varepsilon_p,
&
\left|\delta_\pi(\psi_t,\psi_g)\right|
&\leq \varepsilon_\psi,
\\
|v_{x,t}|
&\leq \varepsilon_v,
&
\left|\delta_\pi(\gamma_t,\gamma_g)\right|
&\leq \varepsilon_\gamma .
\end{aligned}
\label{eq:handover}
\end{equation}
Here $\varepsilon_p$, $\varepsilon_\psi$, $\varepsilon_v$, and $\varepsilon_\gamma$ denote the allowable position, heading, longitudinal-velocity, and articulation errors, respectively. No lateral-velocity condition is imposed for stage completion.

Once the terminal vicinity is reached, the next stage is initialized from the current measured vehicle state rather than from the nominal terminal state of the preceding plan. This prevents residual tracking error from being propagated as an assumed initial condition. Stages~2 and~3 track the reverse and forward portions of the same dual-stage plan, so no replanning is performed at the switchback.

At each control update, the local frame, moving reference, and effective input matrix $B_t$ are updated before solving \eqref{eq:koopman_mpc_problem}; only the first optimal input is applied. The controller uses $\Delta t=0.05$~s, $N=60$ ($T_h=3.0$~s), $u_e\in[-0.5,0.8]$, $u_\gamma\in[-0.3,0.3]$~rad/s, $\gamma\in[-35^\circ,35^\circ]$. OSQP \cite{stellato2020osqp} is used to solve the resulting quadratic programs. Stage-specific parameters are summarized in Table~\ref{tab:mpc_parameters}.

\section{Results and Discussion}
\label{sec:results}

The developed framework is evaluated in three stages: Koopman model identification and selection, nonlinear V-cycle planning, and closed-loop bilinear Koopman MPC execution in the high-fidelity simulation.

\subsection{Deep Bilinear Koopman Model}
\label{subsec:koopman_results}

Thirty Optuna \cite{akiba2019optuna} trials were conducted separately for the forward and reverse models. The reverse model from trial~16 achieved the lowest selection objective and was deployed directly with $n_\phi=32$ and $n_z=38$. For forward motion, trial~20 achieved the lowest objective with $n_z=70$, while trial~13 achieved a comparable objective ($0.956$ versus $0.931$) and terminal-position RMSE ($0.351$ versus $0.339$~m) using only $n_z=38$. Trial~13 was therefore retained as the compact forward candidate. A separate computational-sensitivity study evaluated the effect of lifted dimension and prediction horizon on MPC solution time using $6000$ OSQP solves. At the deployed $3$-s horizon, the selected $n_z=38$ forward and reverse models achieved mean solution times of $24.0$ and $27.2$~ms, respectively. In comparison, the $n_z=70$ forward model required $57.7$~ms on average and exceeded the $50$-ms control period in $89.5\%$ of solves. $n_z=38$ was selected for both deployed models. All timing tests were performed on an Intel Core i9-14900F processor.

Open-loop prediction accuracy was also evaluated using all $300$ held-out test episodes in each direction. For each episode, the full bilinear model and the fixed-state linearized model used by the MPC were initialized from the same measured state and propagated over the deployed $3$-s prediction horizon using the recorded input sequence. For forward motion, the full bilinear model achieved position, heading, $v_x$, $v_y$, and articulation RMSEs of $0.128$~m, $0.566^\circ$, $0.121$~m/s, $0.015$~m/s, and $0.210^\circ$, respectively. The corresponding errors for the fixed-state approximation were $1.327$~m, $2.187^\circ$, $0.694$~m/s, $0.043$~m/s, and $0.324^\circ$. For reverse motion, the full bilinear model achieved $0.148$~m, $0.474^\circ$, $0.123$~m/s, $0.018$~m/s, and $0.172^\circ$, whereas the fixed-state approximation produced $1.187$~m, $3.319^\circ$, $0.576$~m/s, $0.035$~m/s, and $0.505^\circ$, respectively.

Figure~\ref{fig:koopman_openloop} shows an example forward-motion rollout extended to $5$~s (the training tensor horizon). The full bilinear rollout remains close to the AGX trajectory, whereas the fixed-state approximation gradually diverges as the lifted state moves away from the state at which the bilinear coupling was frozen beyond 3 seconds. The vertical dashed line marks the deployed $3$-s MPC horizon.

\subsection{V-Cycle Planning and Closed-Loop Koopman MPC Results}
\label{subsec:vcycle_results}

\begin{figure*}[t]
\centering
\includegraphics[width=\textwidth]{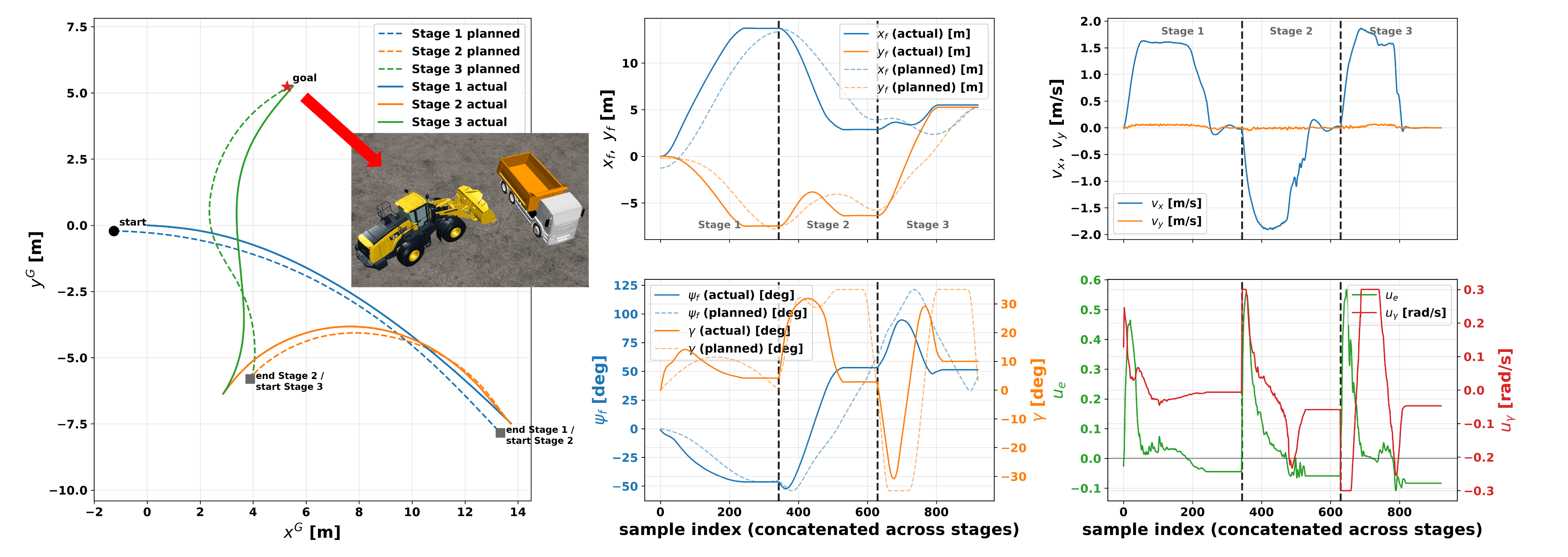}
\caption{Closed-loop execution of the three-stage V-cycle. Left: nonlinear-planner trajectories (dashed) and corresponding trajectories executed by the high-fidelity AGX WA475 plant under Koopman MPC (solid). Center: planned and measured front-axle position, heading, and articulation. Right: measured longitudinal and lateral velocities and the corresponding effort and articulation-rate commands. Vertical dashed lines indicate stage transitions.}
\label{fig:vcycle_closed_loop_results}
\end{figure*}

The complete three-stage V-cycle was evaluated in a continuous simulation. Figure~\ref{fig:vcycle_closed_loop_results} shows the planned and executed trajectories, state evolution, and control inputs, while Table~\ref{tab:vcycle_results} summarizes the performance and computational results.

\begin{table}[t]
\centering
\caption{Planning and closed-loop V-cycle results.}
\label{tab:vcycle_results}
\footnotesize
\setlength{\tabcolsep}{2.5pt}
\begin{tabular}{lccc}
\hline
\textbf{Metric} & \textbf{Stage 1} & \textbf{Stage 2} & \textbf{Stage 3} \\
& \textbf{Forward} & \textbf{Reverse} & \textbf{Forward} \\
\hline
Path length [m]                 & 17.015 & 11.554 & 11.734 \\
Planner time [ms]               & 457.3 & \multicolumn{2}{c}{859.3 (joint)} \\
Final position error [m]        & 0.525 & 1.164 & 0.221 \\
Final heading error [deg]       & 1.363 & 10.985 & 6.325 \\
Final articulation error [deg]  & 4.174 & 5.170 & 9.993 \\
Mean MPC time [ms]              & 23.89 & 25.31 & 25.72 \\
\hline
\end{tabular}
\end{table}

Stage~1 is planned independently in $457.3$~ms, while Stages~2 and~3 are obtained from a single coupled nonlinear optimization in $859.3$~ms, which jointly determines the switchback configuration and truck-approach trajectory. These optimizations are performed only once when each long-horizon reference is generated. The planner provides geometric rather than time-synchronized guidance: the MPC reference is selected from the vehicle's current arclength position along the path, so temporal lead or lag does not accumulate as a tracking error. This allows the reduced-order planner to determine the global maneuver geometry while the Koopman MPC locally realizes it using the learned higher-fidelity dynamics. This behavior is particularly evident during the truck-approach stage. The geometric plan reaches the imposed $\pm35^\circ$ articulation bound, whereas the closed-loop vehicle approaches the truck with a less aggressive articulation profile. Because trajectory tracking and the terminal configuration are enforced through soft costs and terminal-vicinity tolerances, exact reproduction of the kinematic plan is unnecessary once the operational objective is satisfied. Stage~3 reaches a final position error of $0.221$~m and heading error of $6.325^\circ$ without reproducing the full planned articulation excursion. The simulation snapshot in Fig.~\ref{fig:vcycle_closed_loop_results} illustrates that, despite the remaining heading offset, the loader reaches a practically suitable configuration relative to the truck, avoiding high steering near the terminal position.

All three stages reached their configured terminal vicinities while satisfying the imposed state and input bounds. Across the complete maneuver, $616$ MPC problems were solved with zero solver failures. Mean solution times remained near $25$~ms, with a maximum of $35.02$~ms, and no solve exceeded the $50$-ms control period, demonstrating real-time closed-loop execution over the complete V-cycle.

\section{Conclusion}
This work presented a hierarchical model-based-data-driven framework for automating wheel-loader V-cycle maneuvers. The proposed architecture combines a high-level path planner based on a reduced-order articulated-wheel-loader kinematic model with a computationally efficient low-level MPC trajectory-tracking controller. The path planner jointly optimizes the forward and reverse segments of the maneuver and generates geometrically consistent reference trajectories for the low-level controller. To capture the nonlinear vehicle dynamics and vehicle–terrain interactions, the MPC formulation employs deep bilinear Koopman models that learn separate high-dimensional bilinear representations for forward and reverse motions. Data generation and model validation were conducted using a high-fidelity wheel-loader simulation in Algoryx Dynamics.

The simulation results demonstrate the effectiveness of the proposed framework for autonomous execution of V-cycle maneuvers. The proposed Koopman-based MPC enables real-time implementation by deploying a quadratic-programming formulation. The mean MPC computation time remained below 30 ms, well within the 50-ms control sampling interval, and no backup controller was required during the evaluated maneuvers. These results demonstrate the potential of the proposed framework as a computationally efficient approach to V-cycle automation and provide a foundation for extending data-driven predictive control to broader automation tasks involving articulated construction machinery.
\bibliography{references}

\end{document}